\documentclass[letterpaper]{article} 
\usepackage{aaai2026}  
\usepackage{times}  
\usepackage{helvet}  
\usepackage{courier}  
\usepackage[hyphens]{url}  
\usepackage{graphicx} 
\usepackage{natbib}  
\usepackage{caption} 
\usepackage{algorithm}
\usepackage{algorithmic}

\usepackage{newfloat}
\usepackage{listings}
\DeclareCaptionStyle{ruled}{labelfont=normalfont,labelsep=colon,strut=off} 
\floatstyle{ruled}
\newfloat{listing}{tb}{lst}{}
\floatname{listing}{Listing}
\title{RAW-Flow: Advancing RGB-to-RAW Image Reconstruction with Deterministic Latent Flow Matching}
\author{
    Zhen Liu\textsuperscript{\rm 1,\equalcontrib},
    Diedong Feng\textsuperscript{\rm 1,\equalcontrib},
    Hai Jiang\textsuperscript{\rm 2},
    Liaoyuan Zeng\textsuperscript{\rm 1,\footnotemark[2]}, \\
    Hao Wang\textsuperscript{\rm 3},
    Chaoyu Feng\textsuperscript{\rm 3},
    Lei Lei\textsuperscript{\rm 3},
    Bing Zeng\textsuperscript{\rm 1},
    Shuaicheng Liu\textsuperscript{\rm 1,\footnote{Corresponding authors}}
}
\affiliations{
    \textsuperscript{\rm 1}University of Electronic Science and Technology of China \\
    \textsuperscript{\rm 2}Sichuan University \\
    \textsuperscript{\rm 3}Independent Researcher \\
    \{liuzhen03@std., fengdiedong@std., lyzeng@, eezeng@, liushuaicheng@\}uestc.edu.cn, jianghai@stu.scu.edu.cn
}

\usepackage{bibentry}

\usepackage{booktabs}
\usepackage{multirow}
\usepackage{multicol}
\usepackage{enumerate}
\usepackage{array}

\usepackage{amsmath,amsfonts}
\usepackage{algorithmic}
\usepackage{algorithm}
\usepackage{array}
\usepackage{textcomp}
\usepackage{url}
\usepackage{verbatim}
\usepackage{graphicx}
\usepackage{cite}
\usepackage{bbding}
\usepackage{color}

\newcommand{\etal}{\emph{et al.}}

\newcommand{\subtitle}[1]{{\noindent}{\textbf{#1}}}

\begin{document}

\maketitle

\begin{abstract}
RGB-to-RAW reconstruction, or the reverse modeling of a camera Image Signal Processing (ISP) pipeline, aims to recover high-fidelity RAW data from RGB images. Despite notable progress, existing learning-based methods typically treat this task as a direct regression objective and struggle with detail inconsistency and color deviation, due to the ill-posed nature of inverse ISP and the inherent information loss in quantized RGB images. To address these limitations, we pioneer a generative perspective by reformulating RGB-to-RAW reconstruction as a deterministic latent transport problem and introduce a novel framework named \textbf{RAW-Flow}, which leverages flow matching to learn a deterministic vector field in latent space, to effectively bridge the gap between RGB and RAW representations and enable accurate reconstruction of structural details and color information. To further enhance latent transport, we introduce a cross-scale context guidance module that injects hierarchical RGB features into the flow estimation process. Moreover, we design a dual-domain latent autoencoder with a feature alignment constraint to support the proposed latent transport framework, which jointly encodes RGB and RAW inputs while promoting stable training and high-fidelity reconstruction. Extensive experiments demonstrate that RAW-Flow outperforms state-of-the-art approaches both quantitatively and visually. 

\end{abstract}

\begin{links}
    \link{Code}{https://github.com/liuzhen03/RAW-Flow}
\end{links}

\section{Introduction}

Compared to compressed sRGB images that have undergone in-camera Image Signal Processing (ISP), RAW images, which retain the original linear irradiance measurements captured by the camera sensor, typically offer higher bit-depth and richer scene information, thereby being more amenable to accurate modeling by learning-based networks. A growing body of research has demonstrated the advantages of RAW data across various computer vision tasks, including object detection~\cite{raw_detection}, low-light image enhancement~\cite{diffuseraw, SIED}, image super-resolution~\cite{Rbsformer}, and image denoising~\cite{raw_denoising}. However, collecting and annotating large-scale sensor-specific RAW datasets is prohibitively costly, and the large spatial dimension of RAW files places enormous strain on data storage and transmission. To mitigate these limitations, researchers have actively explored reconstructing RAW images from readily available RGB inputs, commonly known as \textit{RGB-to-RAW reconstruction}.
\begin{figure}[t]
    \centering
    \includegraphics[width=1.0\linewidth]{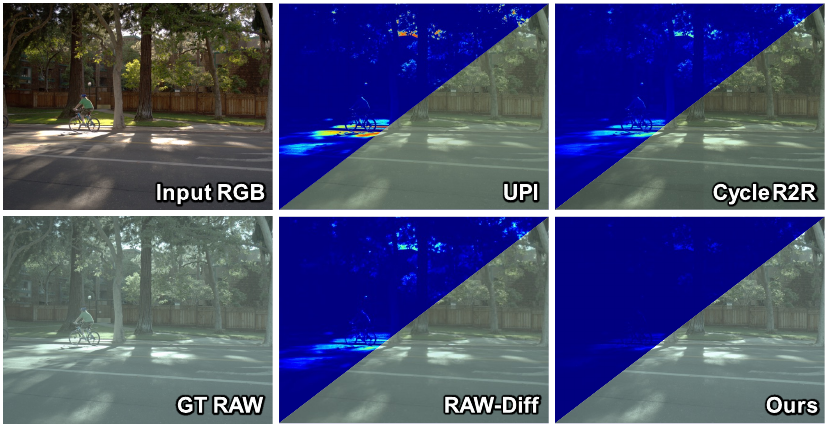}
    \caption{Visual comparisons with previous state-of-the-art methods. The proposed RAW-Flow reconstructs higher-fidelity local details, global luminance, and color information. Differences are best observed in the error maps.}
    \label{fig: teaser}
\end{figure}

Existing learning-based RGB-to-RAW reconstruction approaches can be broadly grouped into two categories. The first class~\cite{CycleR2R,conde2022model} incorporates additional priors or metadata (e.g., exposure, white balance) to approximate individual components of the ISP pipeline. While physically grounded, such methods are prone to cumulative reconstruction errors due to stage-wise modeling and often depend on metadata that is unavailable in real-world scenarios. The second category of methods~\cite{InvISP, reraw} learns a direct, camera-specific mapping from RGB to RAW using neural networks, thus attracting increased attention. Although these models bypass explicit priors dependency, the inversion of ISP remains inherently ill-posed, especially when attempting to reconstruct high-fidelity RAW signals from quantized, clipped RGB images. Nevertheless, despite employing increasingly complex architectures and training objectives, these methods still struggle with generalization, color ambiguity, and incomplete detail recovery. As shown in Fig.~\ref{fig: teaser}, previous methods such as UPI~\cite{UPI} and CycleR2R~\cite{CycleR2R} exhibit color shifts and blurred textures, resulting in degraded reconstruction quality.

Recently, generative models, particularly diffusion models~\cite{ddpm,ddim}, have achieved impressive success in image synthesis and restoration tasks~\cite{ldm, DDRM,sr_diffusion}, owing to their strong expressiveness and capacity to model complex data distributions. However, applying diffusion models directly to RGB-to-RAW reconstruction poses unique challenges. Specifically, learning a deterministic and semantically aligned mapping through noisy intermediate states is difficult due to the inherently ill-posed nature of ISP. This ambiguity often forces the model to over-rely on degraded or semantically mismatched RGB guidance, thereby compromising reconstruction fidelity. As shown in Fig.\ref{fig: teaser}, the recently published diffusion-based RAW-Diff\cite{rawdiffusion}, which uses encoded RGB features for guidance, generates RAW images with noticeable structural distortions. Moreover, diffusion models are notoriously computationally expensive, often requiring hundreds of iterative denoising steps to converge, which limits their practical applicability in RAW reconstruction tasks.

In this paper, we address the aforementioned challenges by leveraging flow matching~\cite{lipman2022flow}, an alternative generative paradigm that learns time-dependent vector fields under direct supervision. Compared to diffusion models, flow matching enables more efficient and stable training by avoiding stochastic denoising. More specifically, we introduce RAW-Flow, a novel deterministic latent flow matching framework for high-fidelity RGB-to-RAW reconstruction. Unlike prior works that operate in pixel space, our approach formulates reconstruction as a latent-space transport problem, where a deterministic vector field is learned to move interpolated latent representations from the RGB domain to the RAW domain. To enhance the accuracy of latent transport, we propose a cross-scale context guidance module that injects hierarchical RGB features into the flow estimation process, promoting improved structural fidelity and chromatic consistency. To support the latent framework, we further design a Dual-domain Latent Autoencoder (DLAE) that jointly encodes RGB and RAW images into their respective latent spaces while ensuring high-quality reconstruction in both domains. A dual-domain feature alignment loss is introduced to stabilize training by encouraging early-layer features in the RAW encoder to align with those from the RGB encoder. As shown in Fig.~\ref{fig: teaser}, the proposed RAW-Flow framework enables faithful reconstruction of sensor-level RAW signals from RGB inputs. In summary, the main contributions of this work are as follows:

\begin{itemize}
    \item We propose a novel deterministic latent flow matching framework, termed RAW-Flow, for RGB-to-RAW reconstruction, which models latent transport via a supervised vector field and incorporates a cross-scale context guidance module to improve luminance and detail fidelity.
    \item We introduce a Dual-domain Latent Autoencoder (DLAE) to jointly encode and reconstruct RGB and RAW images. A dual-domain feature alignment constraint is employed to stabilize RAW learning by aligning shallow features with their RGB counterparts.
    \item Extensive experiments demonstrate that our method outperforms existing state-of-the-art approaches and is capable of reconstructing high-fidelity RAW images.
    
\end{itemize}

\section{Related Work}
\subsection{RGB-to-RAW Reconstruction}
Reconstructing RAW data from RGB images has been a long-standing challenge in low-level vision. Traditional approaches~\cite{chakrabarti2009empirical,debevec2023recovering,chakrabarti2014modeling,grossberg2003determining,lin2005determining,mitsunaga1999radiometric} typically address this problem by calibrating the radiometric response function of cameras through controlled multi-exposure. For example, Mitsunaga~\etal~\cite{mitsunaga1999radiometric} fit a polynomial model to the response curve and iteratively refine it, thereby avoiding the need for precise exposure values. Debevec~\etal~\cite{debevec2023recovering} use a set of images with varying exposures and a smoothness prior to predict a non-parametric inverse response function. However, these methods require careful per-camera calibration and controlled image capture, limiting their applicability to real-world scenes.

Recently, deep learning-based approaches have substantially advanced RGB-to-RAW reconstruction, which can be broadly divided into two categories. The first line of works~\cite{UPI, CycleISP, conde2022model, CycleR2R} seek to explicitly invert individual or grouped components of the ISP pipeline. For instance, UPI~\cite{UPI} decomposes the forward ISP into modular steps and inverts them using camera-specific priors to synthesize realistic RAW data. However, these methods often rely on ISP priors that are rarely available, and their decoupled modeling can lead to cumulative reconstruction errors. The second class~\cite{InvISP,reraw,rawdiffusion} bypasses explicit ISP decomposition and instead learns an end-to-end RGB-to-RAW mapping. InvISP~\cite{InvISP} leverages normalizing flows to model the ISP as an invertible neural operator. ReRAW~\etal~\cite{reraw} employ a multi-head ensemble to generate multiple RAW hypotheses, capturing diverse sensor characteristics. RAW-Diff~\etal~\cite{rawdiffusion} further explores diffusion-based generative modeling to reconstruct RAW images from RGB conditions.
\begin{figure*}[t]
    \centering
    \includegraphics[width=1.0\linewidth]{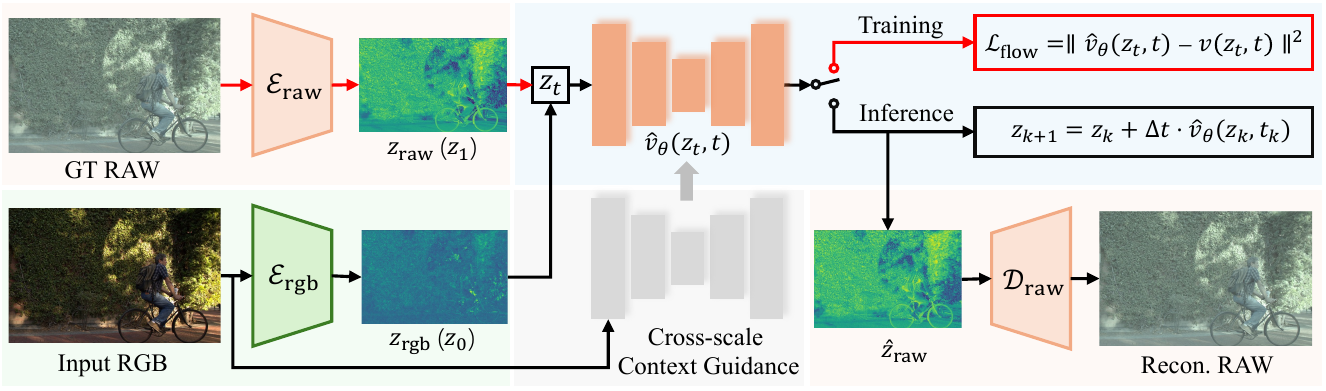}
    \caption{The overall pipeline of our proposed RAW-Flow framework. Given an input RGB image, the RGB encoder $\mathcal{E}_{\text{rgb}}$ extracts the initial latent $\mathbf{z}_{\text{rgb}}$ ($\mathbf{z}_0$), while the RAW encoder $\mathcal{E}_{\text{raw}}$ provides the target latent $\mathbf{z}_{\text{raw}}$ ($\mathbf{z}_1$) during training. A deterministic vector field $\hat{\mathbf{v}}_\theta(\mathbf{z}_t, t)$ is learned to model the latent flow between $\mathbf{z}_0$ and $\mathbf{z}_1$, with cross-scale context guidance injected to enhance the flow estimation. During inference, the predicted flow guides the transport of RGB latent features toward the RAW domain. The resulting latent $\hat{\mathbf{z}}_{\text{raw}}$ is then decoded by $\mathcal{D}_{\text{raw}}$ to reconstruct the RAW image.}
    \label{fig:pipeline}
\end{figure*}

\subsection{Flow Matching}
Flow Matching (FM)~\cite{lipman2022flow} is a recently emerging family of generative models inspired by optimal transport theory~\cite{mozaffari2016optimal}. It learns a vector field that defines an ordinary differential equation (ODE), whose solution maps samples from a prior distribution to the target distribution. Compared to diffusion models such as DDPM~\cite{ddpm} and DDIM~\cite{ddim}, FM allows direct supervision of the vector field and avoids the cumbersome forward and reverse sampling process, resulting in simpler and more numerically stable training. Owing to its lower computational overhead and stronger capacity, FM has been applied across various domains and has demonstrated strong performance, including in image generation~\cite{esser2024scaling}, image restoration~\cite{martin2024pnp}, depth estimation~\cite{depthfm}, and robotics~\cite{hu2024adaflow,zhang2025flowpolicy}. In this work, we formulate the RGB-to-RAW reconstruction task as a latent-space transport problem, where we construct a deterministic vector field that maps interpolated latent features from RGB to RAW representations. 

\section{Method}

\subsection{Overview}
The overall pipeline of our proposed RAW-Flow is illustrated in Fig.~\ref{fig:pipeline}. Given an input RGB image $I_{\text{rgb}}$, we first encode it into a latent representation $\mathbf{z}_{\text{rgb}}$ using an RGB encoder $\mathcal{E}_{\text{rgb}}$. The proposed deterministic latent flow matching module then learns a transport path that maps the RGB latent distribution toward the RAW latent distribution. The resulting raw latent $\hat{\mathbf{z}}_{\text{raw}}$ is subsequently decoded by a RAW decoder $\mathcal{D}_{\text{raw}}$ to reconstruct the corresponding RAW image $\hat{I}_{\text{raw}}$. In the following subsections, we first introduce the proposed Dual-domain Latent Autoencoder, which learns to encode RGB and RAW images into their respective latent representations while enabling high-fidelity reconstruction in both domains. We then describe the deterministic latent flow matching that bridges the RGB and RAW latent spaces. Finally, we detail the training strategy that ensures stable and effective optimization of the entire framework.

\subsection{Dual-domain Latent Autoencoder}
To enable latent-space modeling of the RGB-to-RAW mapping, we first construct high-quality autoencoders for both the RGB and RAW domains. Each autoencoder learns to encode an input image into a compact latent representation from which a high-fidelity reconstruction can be obtained. However, we observe that training an autoencoder in the RAW domain is significantly less stable than in the RGB domain, where directly minimizing pixel-wise reconstruction loss often fails to yield perceptually accurate RAW outputs.
\begin{figure}[!t]
    \centering
    \includegraphics[width=1.0\linewidth]{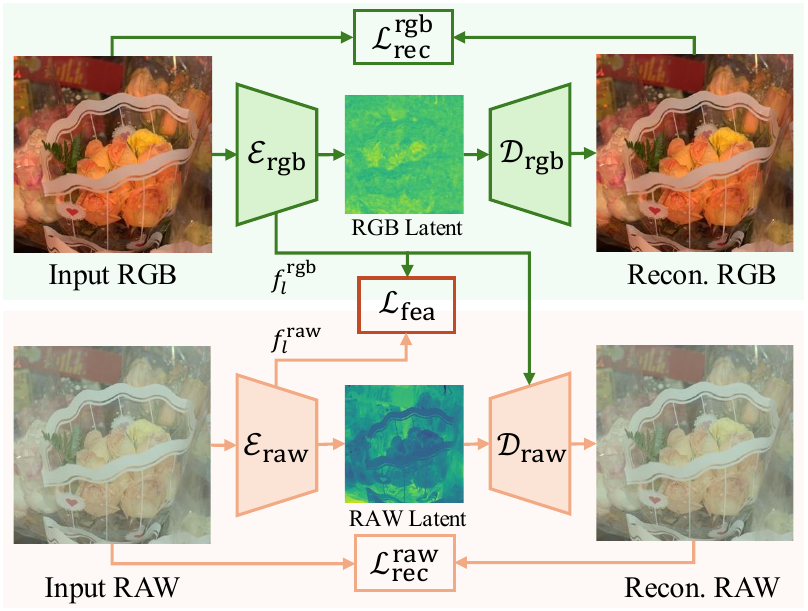}
    \caption{Overview of the Dual-domain Latent Autoencoder (DLAE), which jointly encodes RGB and RAW inputs to enable latent-space alignment and high-fidelity reconstruction.}
    \label{fig:ae_pipeline}
\end{figure}

To this end, we propose the Dual-domain Latent Autoencoder (DLAE), which is designed to jointly achieve stable training and high-quality reconstruction for both RGB and RAW images. As illustrated in Fig.~\ref{fig:ae_pipeline}, the RGB and RAW autoencoders are jointly optimized, where both branches are optimized by a reconstruction loss that combines pixel-wise L2 distance and perceptual similarity. To further stabilize the RAW encoder, we introduce a dual-domain feature alignment loss that encourages the shallow features of the RAW encoder to align with those of the RGB encoder, which serves as a more stable and semantically coherent reference: 
\begin{equation}\label{eq:1}
    \mathcal{L}_{\text{fea}} = \sum_{l \in \mathcal{S}} || f^{\text{raw}}_l - f^{\text{rgb}}_l ||_2^2,
\end{equation}
where $f^{\text{raw}}_l$ and $f^{\text{rgb}}_l$ denote features from layer $l$ in the RAW and RGB encoders, respectively, and $\mathcal{S}$ denotes the set of shallow layers selected for alignment. In addition, the shallow features from the RGB encoder are injected into the corresponding layers of the RAW decoder, where they serve as cross-domain guidance that reinforces both structural detail and color consistency during decoding.

While the RGB and RAW branches are trained alternately, the dual-domain alignment mechanism enables the RAW encoder to benefit from the semantically rich and robust representations learned by the RGB encoder. Although RGB and RAW images differ in format and statistical distribution, they capture the same underlying scene content, motivating the alignment of their low-level feature representations. Meanwhile, deeper latent features are allowed to preserve domain-specific characteristics, as they encode modality-dependent information. Consequently, the proposed DLAE promotes stable optimization and enables high-fidelity reconstruction from RAW latent representations.

\subsection{Deterministic Latent Flow Matching}
After obtaining latent representations from the RGB and RAW autoencoders, we introduce the Deterministic Latent Flow Matching (DLFM) module to approximate the optimal transport from the source RGB latent distribution $\mathbf{z}_0 = \mathcal{E}_{\text{rgb}}(\mathbf{I}_\text{rgb})$ to the target RAW latent distribution $\mathbf{z}_1 = \mathcal{E}_{\text{raw}}(\mathbf{I}_\text{raw})$, where $\mathbf{I}_\text{rgb}$ and $\mathbf{I}_\text{raw}$ denote the paired RGB and RAW images sampled from the dataset, respectively.

The stochastic variants of flow matching typically adopt a training paradigm in which the initial state is sampled from a standard Gaussian distribution, and the intermediate samples are constructed as follows:
\begin{equation}
    \mathbf{z}_t = t \cdot \mathbf{z}_1 + (1 - t) \cdot \boldsymbol{\epsilon}, \quad t \in [0, 1],
\end{equation}
where $t$ denotes the continuous time variable and $\boldsymbol{\epsilon} \sim \mathcal{N}(0, \mathbf{I})$. The model then learns a time-dependent vector field $\hat{\mathbf{v}}_\theta(\mathbf{z}_t, t)$ that guides the flow from noise to data. While effective for unconditional generative modeling, such stochastic sampling introduces unnecessary uncertainty when learning deterministic transformations.

In contrast to the stochastic formulation, we propose a fully deterministic variant tailored for RGB-to-RAW reconstruction, where the mapping between domains is inherently one-to-one under a fixed camera and scene. Instead of sampling noisy initial states, our approach directly models a latent transport path between the RGB and RAW representations, resulting in more stable training and precise latent reconstruction. Specifically, the intermediate latent representation in our DLFM is defined as:
\begin{equation}\label{eq:2}
\mathbf{z}_t = t \cdot \mathbf{z}_1 + (1 - t) \cdot \mathbf{z}_0, \quad t \in [0, 1],
\end{equation}
where a neural network is optimized to predict a time-dependent velocity field $\hat{\mathbf{v}}_\theta(\mathbf{z}_t, t)$ that matches the target velocity vector $\mathbf{v}(\mathbf{z}_t, t) = \mathbf{z}_1 - \mathbf{z}_0$ to represent a constant direction from RGB to RAW latent features. The training objective minimizes the mean squared error between the predicted and ground-truth velocities:
\begin{equation}\label{eq:4}
    \mathcal{L}_{\text{flow}} = \mathbb{E}_{t \sim \mathcal{U}(0,1)} \left\| \hat{\mathbf{v}}_\theta(\mathbf{z}_t, t) - \mathbf{v}(\mathbf{z}_t, t) \right\|^2.
\end{equation}
This deterministic formulation avoids the uncertainty introduced by noise and better aligns with the structural determinism of the RGB-to-RAW task, enabling robust and efficient latent-space transport.

During inference, we recover the RAW latent representation by integrating the learned velocity field starting from the RGB latent $\mathbf{z}_0 = \mathbf{z}_{\text{rgb}}$. Specifically, we uniformly sample $K$ time steps $\{t_k\}_{k=0}^{K-1}$ over $[0, 1]$, and iteratively update the latent representation using Euler integration:
\begin{equation}
    \mathbf{z}_{k+1} = \mathbf{z}_{k} + \Delta t \cdot \hat{\mathbf{v}}_\theta(\mathbf{z}_k, t_k), \quad \text{with } \mathbf{z}_0 = \mathbf{z}_{\text{rgb}},
\end{equation}
where \( \Delta t = \frac{1}{K} \) is the step size. After \( K \) steps, the final transported latent is denoted as \( \hat{\mathbf{z}}_{\text{raw}} = \mathbf{z}_K \). The corresponding RAW image is then reconstructed via the RAW decoder $ \hat{I}_{\text{raw}} = \mathcal{D}_{\text{raw}}(\hat{\mathbf{z}}_{\text{raw}})$. This integration-based inference aligns with the deterministic training objective and allows stable, efficient reconstruction of high-fidelity RAW outputs without the need for stochastic sampling or iterative denoising.

\subtitle{Cross-scale Context Guidance.} Although the DLFM provides a stable path between RGB and RAW latent representations, relying solely on latent embeddings may result in limited awareness of global luminance, color consistency, and fine-grained local structures. To address this limitation, we introduce a cross-scale context guidance module that injects multi-scale context information from the input RGB image into the flow estimation process.

Concretely, we first extract features of the input RGB image through several residual blocks, followed by a U-Net encoder that captures hierarchical representations $\{ \mathbf{f}_i^{\text{rgb}} \}_{i=1}^{L}$ across multiple spatial scales. These hierarchical multi-scale features are subsequently fed into the corresponding layers of the flow matching U-Net, where each scale-specific feature $\mathbf{f}_i^{\text{rgb}}$ provides spatially aligned guidance to its respective matching layer in the flow estimator. Under this design, the time-dependent velocity field becomes conditioned not only on the interpolated latent $\mathbf{z}_t$ and time $t$, but also on the multi-scale context features extracted from the RGB input:
\begin{equation}
    \hat{\mathbf{v}}_\theta(\mathbf{z}_t, t \mid \{\mathbf{f}_i^{\text{rgb}}\}_{i=1}^{L}).
\end{equation}
Compared to single-scale or shallow feature injection, our cross-scale guidance enriches the flow estimation with both global semantics and local textures, enabling more precise alignment between RGB and RAW latent spaces.

\begin{table*}[!t]
\centering
\resizebox{1.0\linewidth}{!}{
\begin{tabular}{
    >{\raggedright\arraybackslash}p{1.8cm}  
    | >{\centering\arraybackslash}p{1.6cm}  
    | >{\centering\arraybackslash}p{1cm} >{\centering\arraybackslash}p{1cm}
    >{\centering\arraybackslash}p{1cm} >{\centering\arraybackslash}p{1cm}
    | >{\centering\arraybackslash}p{1cm} >{\centering\arraybackslash}p{1cm}
    >{\centering\arraybackslash}p{1cm} >{\centering\arraybackslash}p{1cm}
    | >{\centering\arraybackslash}p{1cm} >{\centering\arraybackslash}p{1cm}
    >{\centering\arraybackslash}p{1cm} >{\centering\arraybackslash}p{1cm}
}
\toprule
\multirow{2}{*}{Method} & \multirow{2}{*}{Reference} 
& \multicolumn{4}{c|}{\textbf{FiveK-Nikon}} 
& \multicolumn{4}{c|}{\textbf{FiveK-Canon}} 
& \multicolumn{4}{c}{\textbf{PASCALRAW}} \\
\cmidrule(lr){3-6} \cmidrule(lr){7-10} \cmidrule(lr){11-14}
& & PSNR$_{\text{raw}}$ & SSIM$_{\text{raw}}$ & PSNR$_{\text{rgb}}$ & SSIM$_{\text{rgb}}$
& PSNR$_{\text{raw}}$ & SSIM$_{\text{raw}}$ & PSNR$_{\text{rgb}}$ & SSIM$_{\text{rgb}}$
& PSNR$_{\text{raw}}$ & SSIM$_{\text{raw}}$ & PSNR$_{\text{rgb}}$ & SSIM$_{\text{rgb}}$ \\
\midrule
UNet       & MICCAI'15  & 25.65 & 0.8535 & 25.94 & 0.8583 & 29.20 & 0.9221 & 29.69 & 0.9277 & 33.87 & 0.9645 & 34.43 & 0.9700 \\
UPI        & CVPR'19    & 25.55 & 0.8416 & 25.79 & 0.8479 & 27.24 & 0.8846 & 27.46 & 0.8892 & 26.26 & 0.8828 & 26.50 & 0.8884 \\
CycleISP   & CVPR'20    & 26.41 & 0.8574 & 26.64 & 0.8610 & 29.96 & 0.9389 & 30.42 & 0.9426 & 34.89 & 0.9735 & 35.59 & 0.9794 \\
InvISP     & CVPR'21    & 26.94 & 0.8268 & 27.34 & 0.8319 & 28.32 & 0.8648 & 28.74 & 0.8680 & 28.47 & 0.8631 & 28.80 & 0.8602 \\
CycleR2R   & TPAMI'24   & 25.43 & 0.8272 & 25.52 & 0.8260 & 26.33 & 0.8579 & 26.55 & 0.8588 & 26.19 & 0.8349 & 26.25 & 0.8311 \\
ReRAW-R    & CVPR'25    & 27.90 & 0.8447 & \underline{28.29} & 0.8495 & 30.98 & 0.9230 & 31.34 & 0.9259 & 35.96 & \underline{0.9834} & 36.17 & 0.9837  \\
ReRAW-S    & CVPR'25    & \underline{28.04} & 0.8377 & 28.28 & 0.8413 & 31.76 & 0.9328 & 32.10 & 0.9355 & \underline{37.29} & \textbf{0.9854} & \underline{37.47} & \underline{0.9857} \\
\midrule
DDPM       & NeurIPS'20 & 25.82 & 0.8392 & 26.27 & 0.8435 & 28.70 & 0.8485 & 29.53 & 0.8681 & 32.24 & 0.9574 & 32.87 & 0.9632  \\
RAW-Diff   & WACV'25    & 26.96 & \underline{0.8608} & 27.46 & \underline{0.8653} & \underline{31.84} & \textbf{0.9571} & \underline{32.41} & \textbf{0.9613} & 29.54 & 0.9340 & 30.11 & 0.9395 \\
\midrule
Ours       & -          & \textbf{30.79} & \textbf{0.8772} & \textbf{31.28} & \textbf{0.8816} & \textbf{32.55} & \underline{0.9445} & \textbf{33.30} & \underline{0.9508} & \textbf{37.62} & 0.9831 & \textbf{38.69} & \textbf{0.9892} \\
\bottomrule
\end{tabular}}
\caption{Quantitative comparisons on the FiveK-Nikon, FiveK-Canon, and PASCALRAW test sets. We report PSNR and SSIM for both the reconstructed RAW images and their corresponding converted RGB outputs. The best results are highlighted in \textbf{bold}, and the second-best are \underline{underlined}.}
\label{tab:quantitative_comparison}
\end{table*}

\subsection{Training Strategy}
Our RAW-Flow framework is optimized in two stages to ensure stable convergence and effective representation learning. In the first stage, we train the proposed Dual-domain Latent Autoencoder (DLAE) using paired RGB and RAW images. The training objective of the RGB autoencoder is to minimize a reconstruction loss that combines pixel-wise fidelity and perceptual similarity:
\begin{equation}
    \mathcal{L}_{\text{rec}}^{\text{rgb}} = || \tilde{I}_{\text{rgb}} - I_{\text{rgb}} ||_2^2 + \lambda_1 \cdot || \phi(\tilde{I}_{\text{rgb}}) - \phi(I_{\text{rgb}}) ||_2^2,
\end{equation}
where $\tilde{I}_\text{rgb}$ denotes the reconstructed RGB image from the autoencoder, and $I_\text{rgb}$ is the corresponding ground truth. $\phi(\cdot)$ represents features extracted from a pretrained VGG network. Similarly, the RAW reconstruction loss $\mathcal{L}_{\text{rec}}^{\text{raw}}$ adopts the same formulation as $\mathcal{L}_{\text{rec}}^{\text{rgb}}$ but is applied to the RAW domain (using $I_{\text{raw}}$ and its reconstruction $\tilde{I}_{\text{raw}}$). For the RAW autoencoder, we adopt the proposed $\mathcal{L}_{\text{fea}}$ (i.e., Eq.(\ref{eq:1})) to stabilize training, resulting in the objective as follows:
\begin{equation}
    \mathcal{L}_{\text{raw}} = \mathcal{L}_{\text{rec}}^{\text{raw}} + \lambda_2 \cdot \mathcal{L}_{\text{fea}}.
\end{equation}

In the second stage, we freeze the parameters of the DLAE and train the DLFM module to model the latent transport from RGB to RAW. The training is supervised using the flow loss $\mathcal{L}_{\text{flow}}$ defined in Eq.~\ref{eq:4}, which minimizes the discrepancy between predicted and target velocity vectors across interpolated latent trajectories. Inspired by prior work~\cite{leng2025repae}, which shows that jointly optimizing autoencoders and latent generative models can alleviate representation mismatches and lead to performance gains, we further fine-tune the entire RAW-Flow framework in an end-to-end manner. This allows the autoencoders and DLFM module to co-adapt and better align in the latent space. The objective is defined as:
\begin{equation}
    \mathcal{L} = ||\hat{I}_\text{raw} - I_\text{raw}||_1 + \lambda_3 \cdot ||\phi(\hat{I}_\text{raw}) - \phi(I_\text{raw})||_2^2,
\end{equation}
where $\hat{I}_\text{raw}$ and $I_\text{raw}$ denote the predicted and ground-truth RAW images, respectively. In our experiments, we empirically set $\lambda_1=0.01$, $\lambda_2=0.1$, and $\lambda_3=0.01$.
\begin{figure*}[!t]
    \centering
    \includegraphics[width=1.0\linewidth]{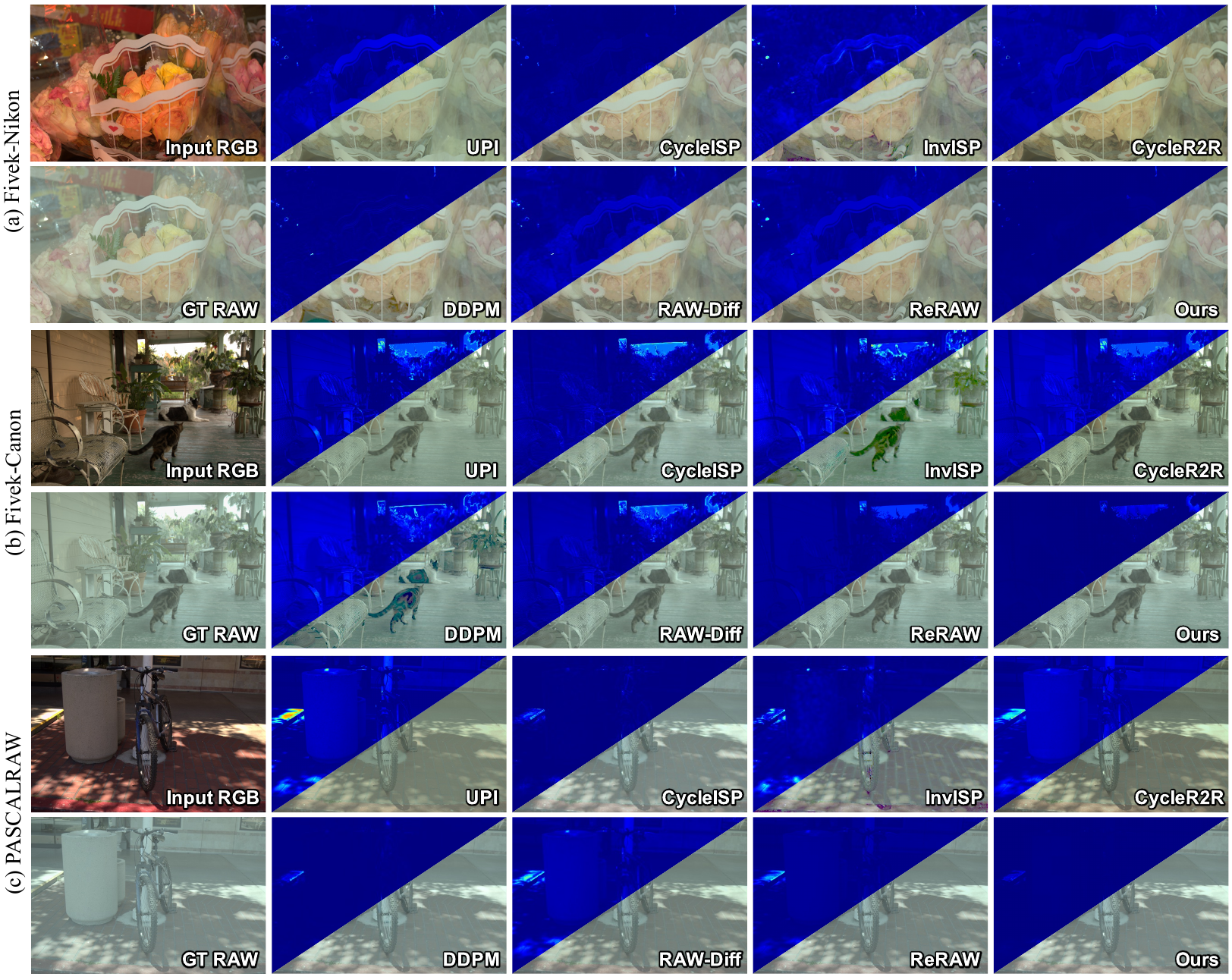}
    \caption{Qualitative comparisons with state-of-the-art RGB-to-RAW reconstruction methods on the FiveK-Nikon, FiveK-Canon, and PASCALRAW datasets. For each method, we show the reconstructed RAW image and the corresponding error map, which visualizes the pixel-wise difference from the ground-truth RAW image (darker regions indicate smaller errors).}
    \label{fig:qualitative_results}
\end{figure*}

\section{Experiments}

\subsection{Experimental Setting}

\subtitle{Datasets.} To evaluate our RGB-to-RAW reconstruction framework, we conduct experiments on three benchmark RAW datasets captured using various DSLR cameras. From the MIT-Adobe FiveK collection~\cite{adobefivek}, we construct two subsets: \textbf{FiveK-Nikon} and \textbf{FiveK-Canon}, containing 590 and 777 RAW images captured by the Nikon D700 and Canon EOS 5D cameras, respectively. Additionally, we construct a subset from the \textbf{PASCALRAW} dataset~\cite{pascalraw}, which was collected using a Nikon D3200 camera. We randomly split the selected data into 85\% for training and 15\% for testing. All RAW images are processed using the RawPy library to generate their corresponding RGB counterparts, resulting in full-resolution RGB-RAW pairs for training and evaluation.

\subtitle{Evaluation Metrics.} Following prior works, we adopt Peak Signal-to-Noise Ratio (PSNR) and Structural Similarity Index (SSIM) to evaluate the quality of the reconstructed RAW images. All metrics are computed at full resolution, and we report the average values over the test split of each dataset.

\subtitle{Implementation Details.} Our framework is implemented using PyTorch and trained with the Adam~\cite{adam} optimizer, with $\beta_1 = 0.9$, $\beta_2 = 0.999$, and $\epsilon = 1\mathrm{e}{-8}$. The initial learning rate for both the DLAE and DLFM is set to $1\mathrm{e}{-4}$, and is decayed by a Reduce-on-Plateau scheduler. The time step K of the DLFM is set to 20. We train the DLAE for 200 epochs and the DLFM for 100 epochs. All experiments are conducted with NVIDIA RTX 4090 GPUs.

\subsection{Comparison with Existing Methods}
\subtitle{Comparison Methods.} To comprehensively evaluate the performance of the proposed RAW-Flow, we compare it against two categories of state-of-the-art methods: 1) regression-based approaches including UNet~\cite{UNet}, UPI~\cite{UPI}, CycleISP~\cite{CycleISP}, InvISP~\cite{InvISP}, CycleR2R~\cite{CycleR2R}, and ReRAW~\cite{reraw}, and 2) diffusion-based models including DDPM~\cite{ddpm} and RAW-Diff~\cite{rawdiffusion}. For fair comparisons, all competing methods are implemented using their official codes and default settings, and are evaluated under the same hardware environment.

\subtitle{Quantitative Comparison.} The quantitative results are summarized in Table~\ref{tab:quantitative_comparison}. It can be observed that the proposed RAW-Flow consistently achieves the best performance across all benchmarks, outperforming both regression-based and diffusion-based methods. Taking FiveK-Nikon as an example, RAW-Flow surpasses the second-best ReRAW by 2.75 dB in RAW-domain PSNR and 3 dB in RGB-domain PSNR. On the other hand, while existing diffusion-based methods generally underperform compared to the best regression-based approach, our flow matching based framework outperforms both groups, demonstrating its effectiveness in learning high-fidelity RAW reconstructions.

\subtitle{Qualitative Comparison.} Fig.~\ref{fig:qualitative_results} presents qualitative comparisons on three representative scenes from the FiveK-Nikon, FiveK-Canon, and PASCALRAW datasets. For better visualization, we include error maps in the top-left corner of each method's result, showing pixel-wise differences from the ground-truth RAW images (darker regions indicate smaller errors). As observed, prior methods often exhibit global luminance or color inconsistencies and struggle to recover fine-grained details, particularly in regions with complex lighting variations and intricate textures. In contrast, the proposed RAW-Flow achieves more faithful reconstructions with enhanced detail preservation and color fidelity.

\begin{table}[!t]
    \centering
    \resizebox{1.0\linewidth}{!}{
    \begin{tabular}{
        >{\centering\arraybackslash}p{0.4cm}  
        >{\centering\arraybackslash}p{0.4cm}  
        >{\centering\arraybackslash}p{0.4cm}  
        | >{\centering\arraybackslash}p{1.2cm} 
        >{\centering\arraybackslash}p{1.2cm}  
        | >{\centering\arraybackslash}p{1.2cm} 
        >{\centering\arraybackslash}p{1.2cm}  
    }
    \toprule
    \multicolumn{3}{c|}{Modules} & \multicolumn{2}{c|}{Autoencoder RAW} & \multicolumn{2}{c}{Reconstructed RAW} \\
    \cmidrule{1-3} \cmidrule{4-5} \cmidrule{6-7}
    $\textbf{z}_\text{raw}$ & $f_\text{rgb}$ & $\mathcal{L}_\text{fea}$ & PSNR $\uparrow$ & SSIM $\uparrow$& PSNR $\uparrow$& SSIM $\uparrow$\\
    \midrule
    \checkmark &               &                & 20.08 & 0.7808 & 29.19 & 0.8605\\
    \checkmark & \checkmark    &                & 25.36 & 0.8579 & 28.48 & 0.8428 \\
    \checkmark & \checkmark    & \checkmark     & \textbf{32.94} & \textbf{0.9895} & \textbf{30.79} & \textbf{0.8772} \\
    \bottomrule
    \end{tabular}}
    \caption{Ablation results of the dual-domain autoencoder (DLAE) design. We analyze the impact of injected RGB features $f_\text{rgb}$ and the feature alignment loss $\mathcal{L}_\text{fea}$ on the RAW autoencoder and the final RAW reconstruction quality.}
    \label{tab:ae_ablation}
\end{table}

\subsection{Ablation Studies}
In this section, we conduct ablation studies to evaluate the contribution of each core component of RAW-Flow.

\begin{figure}[!t]
    \centering
    \includegraphics[width=1.0\linewidth]{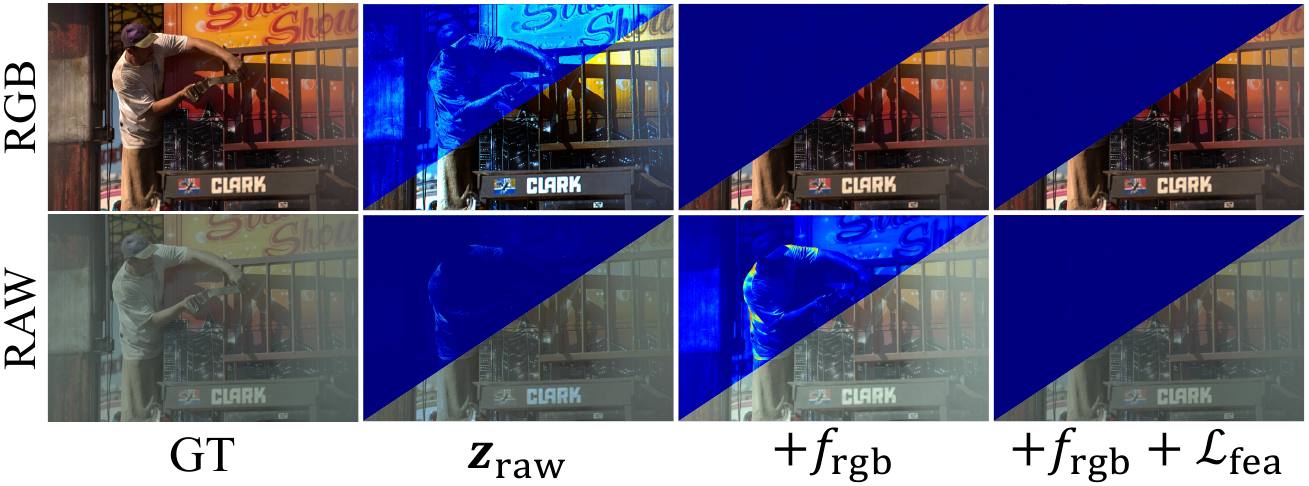}
    \caption{Visual comparisons of our ablation study about the injected feature $f_\text{rgb}$ and the feature alignment loss $\mathcal{L}_\text{fea}$ of the proposed dual-domain autoencoder (DLAE).}
    \label{fig:ablation_ae}
\end{figure}

\subtitle{Dual-domain Latent Autoencoder.} We first analyze the Dual-domain Latent Autoencoder (DLAE), focusing on the effects of the injected RGB features $f_{\text{rgb}}$ and the feature alignment loss $\mathcal{L}_{\text{fea}}$. As shown in Table~\ref{tab:ae_ablation}, incorporating RGB features into the RAW autoencoder yields noticeable improvements in PSNR and SSIM over using the RAW latent $\mathbf{z}_{\text{raw}}$ alone. Nonetheless, directly relying on RGB features for RAW reconstruction remains suboptimal due to the domain gap between RGB and RAW representations. Introducing the feature alignment constraint mitigates this gap by enforcing consistency between the shallow features of the RGB and RAW encoders, while still preserving domain-specific characteristics in the latent space. This leads to further gains in RAW reconstruction fidelity. Visual comparisons in Fig.~\ref{fig:ablation_ae} corroborate these findings, demonstrating that DLAE enables high-quality reconstruction in both domains.

\begin{table}[!t]
\centering
\resizebox{1.0\linewidth}{!}{
\begin{tabular}{l|cc|cc}
    \toprule
    \multirow{2}{*}{Variant} & \multicolumn{2}{c|}{FiveK-Nikon} & \multicolumn{2}{c}{FiveK-Canon} \\
    \cmidrule{2-5}
     & PSNR $\uparrow$ & SSIM $\uparrow$ & PSNR $\uparrow$ & SSIM $\uparrow$ \\
    \midrule
    Diffusion & 28.30 & 0.8235 & 28.79 & 0.8638 \\
    Stochastic FM & 27.32 & 0.8696 & 22.74 & 0.6783 \\
    DLFM (Ours) & \textbf{30.79} & \textbf{0.8772} & \textbf{32.55} & \textbf{0.9445} \\
    \bottomrule
\end{tabular}}
\caption{Quantitative results of ablation studies on the proposed Deterministic Latent Flow Matching (DLFM). We conduct comparisons against a diffusion-based variant and a stochastic flow matching (Stochastic FM) variant.}
\label{tab:fm_ablation}
\end{table}

\begin{table}[!t]
\centering
\resizebox{1.0\linewidth}{!}{
\begin{tabular}{l|cc|cc}
    \toprule
    \multirow{2}{*}{Guidance} & \multicolumn{2}{c|}{FiveK-Nikon} & \multicolumn{2}{c}{FiveK-Canon} \\
    \cmidrule{2-5}
        & PSNR $\uparrow$ & SSIM $\uparrow$ & PSNR $\uparrow$ & SSIM $\uparrow$ \\
    \midrule
    Single-scale Context   & 29.04 & 0.8464 & 28.43 & 0.8587 \\
    RGB Latent             & 28.50 & 0.8712 & 28.37 & 0.9037 \\
    Cross-scale Context    & \textbf{30.79} & \textbf{0.8772} & \textbf{32.55} & \textbf{0.9445} \\
    \bottomrule
\end{tabular}}
\caption{Quantitative results of ablation studies on the proposed Cross-scale Context Guidance. The cross-scale design yields significantly better performance than using only the RGB latent or single-scale context as guidance.}
\label{tab:cscg_ablation}
\end{table}

\begin{figure}[!t]
    \centering
    \includegraphics[width=1.0\linewidth]{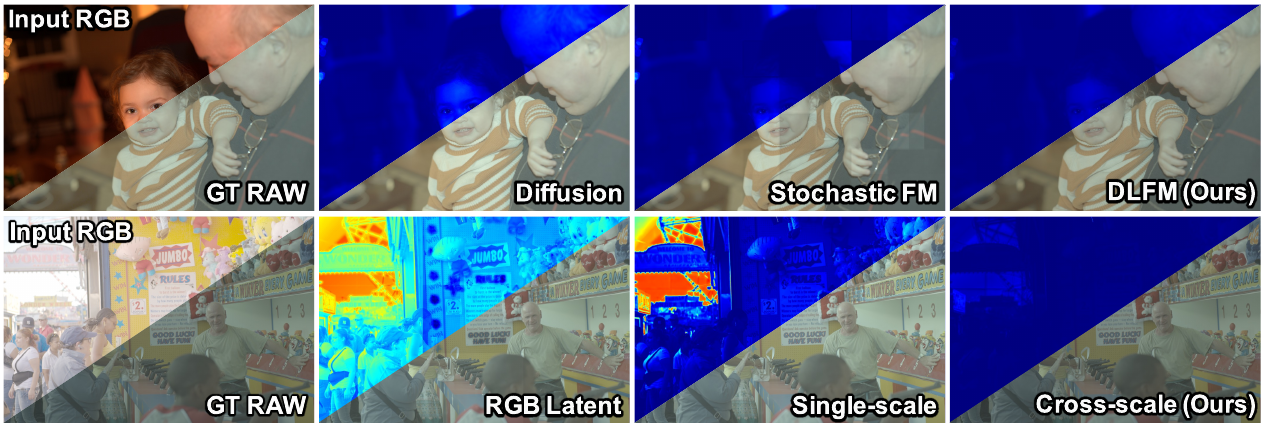}
    \caption{Visual results of our ablation studies on the proposed deterministic latent flow matching (first row) and the cross-scale context guidance mechanism (second row).}
    \label{fig:ablation_guidance}
\end{figure}

\subtitle{Deterministic Latent Flow Matching.} To evaluate the effectiveness of the Deterministic Latent Flow Matching (DLFM), we compare it against two representative alternatives: (1) a standard diffusion model, and (2) a stochastic flow matching variant in which the initial latent representation $\mathbf{z}_0$ is sampled from Gaussian noise. As summarized in Table~\ref{tab:fm_ablation}, DLFM outperforms both baselines by a significant margin, affirming its strength in capturing a deterministic and semantically consistent latent transport path. Visual comparisons in the first row of Fig.~\ref{fig:ablation_guidance} further demonstrate that DLFM yields more accurate and structurally consistent RAW reconstructions, whereas diffusion-based and stochastic counterparts suffer from detail degradation and region-wise inconsistencies due to sampling-induced randomness.

\subtitle{Cross-scale Context Guidance.} We further investigate the contribution of the proposed Cross-scale Context Guidance by replacing it with two alternative forms of guidance: (1) RGB latent features, and (2) single-scale global context extracted by a residual network with comparable parameters. As reported in Table~\ref{tab:cscg_ablation}, both variants lead to noticeable performance degradation. Qualitative comparisons in the second row of Fig.~\ref{fig:ablation_guidance} reveal that relying solely on RGB latent features introduces global luminance shifts. Although single-scale context guidance offers slight improvements in luminance consistency, it still falls short in preserving fine-grained details, particularly in complex image regions. In contrast, the proposed multi-scale guidance enables better luminance consistency and detail fidelity, yielding more faithful and perceptually coherent RAW reconstructions.

\section{Conclusion}
In this work, we have presented RAW-Flow, a novel generative framework for RGB-to-RAW image reconstruction that reformulates the task as a deterministic latent transport problem. Unlike prior regression-based methods, RAW-Flow directly learns a vector field in the latent space via flow matching, effectively bridging the domain gap between RGB and RAW representations. To support this formulation, we design a Dual-domain Latent Autoencoder (DLAE) with a feature alignment constraint to jointly encode RGB and RAW inputs, ensuring stable training and reconstruction fidelity. A cross-scale context guidance module further injects hierarchical RGB information to guide flow estimation. Extensive experiments on multiple benchmarks demonstrate that the proposed RAW-Flow outperforms state-of-the-art approaches both quantitatively and qualitatively.

\section{Acknowledgments}
This work was supported in part by National Natural Science Foundation of China under grant No.62372091 and in part by Hainan Province Science and Technology SpecialFund under grant No. ZDYF2024(LALH)001

\bibliography{aaai2026}

\end{document}